\documentclass{article}
\usepackage{spconf,amsmath,amssymb,graphicx,amsfonts,cite,tabularx,multirow,xcolor,url,tipa,breakurl,bm}
\usepackage[htt]{hyphenat}

\usepackage{xcolor}
\usepackage{amsmath}

\newcolumntype{D}{>{\centering\arraybackslash}m{5.8ex}}
\newcolumntype{E}{>{\arraybackslash}m{19.2ex}}
\newcolumntype{F}{>{\centering\arraybackslash}m{10.5ex}}

\title{Deep Shallow Fusion for RNN-T Personalization}
\name{Duc Le, Gil Keren, Julian Chan, Jay Mahadeokar, Christian Fuegen, Michael L. Seltzer}
\address{Facebook AI\\
	{\small \texttt{\{duchoangle,gilkeren,julianchan,jaym,fuegen,mikeseltzer\}@fb.com}}}
\begin{document}
%
\maketitle
\begin{abstract}
End-to-end models in general, and Recurrent Neural Network Transducer (RNN-T) in particular, have gained significant traction in the automatic speech recognition community in the last few years due to their simplicity, compactness, and excellent performance on generic transcription tasks. However, these models are more challenging to personalize compared to traditional hybrid systems due to the lack of external language models and difficulties in recognizing rare long-tail words, specifically entity names. In this work, we present novel techniques to improve RNN-T's ability to model rare WordPieces, infuse extra information into the encoder, enable the use of alternative graphemic pronunciations, and perform deep fusion with personalized language models for more robust biasing. We show that these combined techniques result in 15.4\%--34.5\% relative Word Error Rate improvement compared to a strong RNN-T baseline which uses shallow fusion and text-to-speech augmentation. Our work helps push the boundary of RNN-T personalization and close the gap with hybrid systems on use cases where biasing and entity recognition are crucial.
\end{abstract}
\section{Introduction}
\label{sec:intro}

End-to-end techniques for automatic speech recognition (ASR), most notably sequence-to-sequence models with attention \cite{Chorowski:2015:AMS:2969239.2969304,chan2015listen,Chiu18,zeyer2018improved} and Recurrent Neural Network Transducer (RNN-T) \cite{Graves12transduction,Prabhavalkar17,Battenberg17RNNT,He2019RNNT}, are becoming increasingly popular. Compared to the traditional hybrid system based on Hidden Markov Model and Deep Neural Network (HMM-DNN) with individually-trained components, all parts of an end-to-end model are optimized jointly, which often leads to better performance on recognition tasks with sufficient training data and low training-testing mismatch. End-to-end systems are simpler to train; they typically do not require pronunciation lexicons, decision trees, initial bootstrapping, nor forced alignment. End-to-end models are also more suitable for on-device use cases due to the lack of external language models (LMs) or decoding graphs, whose sizes can be prohibitively large in hybrid setups because of large vocabulary support, complex LMs, and context-dependent decision trees.

End-to-end systems do have limitations, however. Their end-to-end nature leads to a lack of composability, such as that between acoustic, language, and pronunciation models in hybrid setups. This lack of composability in turn leads to challenges in personalization, which traditionally involves on-the-fly modification of external LMs (or decoding graphs) to add, boost, and penalize certain words or phrases. Previous work in end-to-end ASR addressed this issue by incorporating external LMs during beam search (i.e., shallow fusion), with special modifications to handle the model's spiky output \cite{Kannan2018,Zhao2019,He2019RNNT}. A fundamental limitation of shallow fusion is that it relies on late combination, hence the model needs to have the potential to produce the correct output in the first place without access to biasing information. Another class of method (i.e., deep context) adds an attention-based \cite{Pundak2018DC,Chen2019DC,Jain2020DC} or simple \cite{Jain2020DC} biasing module over contextual phrases to provide additional signal to the decoder component of end-to-end models. While promising, these methods were shown to have problems scaling to large and highly confusable biasing lists.

A closely related challenge of ASR personalization is entity recognition, since in many cases biasing items are entity names. Rare name recognition presents significant challenges to end-to-end models because of two main reasons. First, the output units of end-to-end models are typically graphemes or WordPieces \cite{Irie2019output}, both of which do not work well when the spelling of a word does not correspond to how it is pronounced (e.g., non-English names relative to English spelling conventions). Second, rare names often decompose into target sequences that are not seen enough in training, making them difficult to recognize correctly. By contrast, both problems are alleviated in hybrid systems due to the use of phonetic lexicons and/or clustered context-dependent acoustic targets. Popular solutions to this problem include upsampling entity-heavy data or generating synthetic training data with names using text-to-speech (TTS) \cite{He2019RNNT,Peyser2019,Zhao2019,Rosenberg19ASR+TTS}. While this method alleviates the data sparsity issue, it does not address the underlying problems of under-trained targets and unconventional spelling of rare names. 

In this work, we propose several novel techniques to address both challenges and further improve RNN-T personalization. To alleviate the problem of under-trained targets and recognition of unconventional names, we adopt on-the-fly sub-word regularization \cite{Kudo2018SubWord} to increase WordPiece coverage during training, perform pre-training \cite{Hu2020PreTraining} and multi-task learning (MTL) \cite{Liu2021MTL} to strengthen the encoder, and leverage grapheme-to-grapheme (G2G) \cite{Le2020G2G} to generate alternative graphemic pronunciations for names. To address the limitation of shallow fusion relying on late combination, we introduce deep personalized LM (PLM) fusion to influence the model's predictions earlier. We show that the combination of these techniques results in \textbf{15.4\%}--\textbf{34.5\%} relative Word Error Rate (WER) improvement on top of a strong RNN-T baseline which leverages shallow fusion and TTS augmentation. Our final model is also competitive with a hybrid system that has significantly larger disk and memory footprint.

\section{Background and Related Work}
\label{sec:related_work}

\subsection{Recurrent Neural Network Transducer (RNN-T)}
\label{ssec:rnnt}

The RNN-T model architecture, first proposed in \cite{Graves12transduction}, can be broken down into three different components. Firstly, the \textbf{\emph{encoder}}, which can be thought of as RNN-T's built-in acoustic model (AM), is responsible for transforming a sequence of acoustic feature vectors $\mathbf{x} = (\mathbf{x}_1, \mathbf{x}_2, \ldots, \mathbf{x}_T)$, where $\mathbf{x}_t \in \mathbb{R}^{d}$ and $T$ is the number of frames, into a sequence of high-level representation $\mathbf{h}^{\text{enc}} = (\mathbf{h}_1^{\text{enc}}, \mathbf{h}_2^{\text{enc}}, \ldots, \mathbf{h}_{T'}^{\text{enc}})$:
\begin{equation}
 \mathbf{h}^{\text{enc}} = f^{\text{enc}}(\mathbf{x}) 
\end{equation}
\noindent where $T'$ may be different from $T$ due to subsampling.

Secondly, the \textbf{\emph{predictor}}, which can be thought of as RNN-T's built-in LM, is responsible for transforming a sequence of previous output units $y_1, y_2, \ldots, y_{u-1}$, where $y_u$ is typically a grapheme or WordPiece, into an embedding vector $\mathbf{h}_u^{\text{pred}}$:
\begin{equation}
    \mathbf{h}^{\text{pred}}_u = f^{\text{pred}}(y_1, y_2, \ldots, y_{u-1})
\end{equation}

Thirdly, the \textbf{\emph{joiner}}, which can be thought of as RNN-T's built-in decoder, is responsible for combining the encoder output $\mathbf{h}_t^{\text{enc}}$ at acoustic time step $t$ and predictor output $\mathbf{h}_u^{\text{pred}}$ at prediction time step $u$ to estimate the logits $\mathbf{z}_{t,u}$:
\begin{equation}
    \mathbf{z}_{t,u} = f^{\text{join}}(\mathbf{h}_t^{\text{enc}} + \mathbf{h}_u^{\text{pred}})
\label{eq:joiner}
\end{equation}

Finally, the posterior distribution over the next output symbols at acoustic time step $t$ and prediction time step $u$ can be computed by passing $\mathbf{z}_{t,u}$ through a softmax function:
\begin{equation}
    P(\cdot | \mathbf{x}_1 \ldots \mathbf{x}_t, y_1 \ldots y_{u-1}) = \text{Softmax}(\textbf{z}_{t,u})
\end{equation}

Details about RNN-T's training objective and decoding algorithm can be found in \cite{Graves12transduction}. Compared to other end-to-end architectures, such as sequence-to-sequence models with attention, RNN-T is easier to stream while maintaining highly competitive recognition performance \cite{Battenberg17RNNT,chiu2019comparison,li2020comparison}.

\subsection{External LM Fusion and Deep Context}
\label{ssec:lm_fusion}

One major limitation of RNN-T and end-to-end models in general is that they are reliant on audio training data and cannot easily leverage unpaired text. The most popular method for using text data with RNN-T is via shallow fusion, where RNN-T scores are combined with external LM scores during beam search \cite{Kannan2018,Zhao2019}. Alternatively, external neural network LMs (NNLMs) can be fused during training and augment the end-to-end model's decoder/predictor component with additional linguistic information via cold fusion \cite{gulcehre2015using,sriram2018cold}. The main challenge in applying cold fusion to RNN-T personalization is that we would need to rapidly adapt the NNLM on a very small amount of personalized information, which is difficult to do efficiently and effectively. Deep context \cite{Pundak2018DC,Chen2019DC} can be considered an extension of cold fusion which replaces the NNLM with an attention module over bias words. As shown in \cite{Pundak2018DC}, while this method worked well on smaller biasing lists, it failed to yield improvement on larger lists even after combining with shallow fusion. A recent work replaced the attention module with a non-trainable position-aware biasing module which gave similar results \cite{Jain2020DC}; however, they did not analyze this method's performance on large biasing lists nor compare with shallow fusion baselines.

Our deep PLM fusion technique (Section \ref{ssec:wfst_deep_fusion}) can be viewed as an extension to cold fusion that utilizes a trie-based biasing module which can be constructed efficiently on-the-fly given personalization data. Moreover, our method is able to yield significant improvement with large biasing lists over pure shallow fusion, which previous work did not achieve.

\subsection{Grapheme-to-Grapheme (G2G)}
\label{ssec:g2g}

G2G is a technique originally proposed for hybrid graphemic ASR to improve rare name recognition \cite{Le2020G2G}. The idea is akin to grapheme-to-phoneme (G2P), but the output of the model is a sequence of graphemes instead of phonemes. G2G models are trained by generating artificial data with TTS, re-decoding the data using a specialized graphemic ASR system, and finally training a statistical model to map the original written form to the re-decoded ASR output. Trained G2G models can transform a word into alternative spellings with similar pronunciations, such as ``Kaity" $\rightarrow$ ``Katie." The authors found that using G2G to generate additional pronunciations for decoding led to significant improvement on rare name recognition. In this paper, we extend their work to apply G2G to the end-to-end paradigm. We also investigate using G2G during model training instead of just decoding.

\section{Data}
\label{sec:data}

Our training set is made up of two manually-transcribed anonymized in-house corpora with no personally identifiable information (PII). The first corpus comprises 15.7M utterances (12.5K hours) in the voice assistant domain recorded by 20K crowd-sourced workers on mobile devices. The second corpus contains 1.2M voice commands (1K hours) sampled from the production traffic of Facebook Portal after the ``hey Portal" wakeword is activated. Utterances from this corpus are further morphed when researchers access them, in an effort to de-identify the user. We follow the same data augmentation procedure described in \cite{Le2020G2G}, which involves a mix of simulated Room Impulse Response (RIR), additive background noise extracted from public Facebook videos, and speed perturbation \cite{Ko2015AudioAF}. The final distorted dataset contains 38.6M utterances (31K hours).

Our evaluation set comprises 20.8K manually-transcribed anonymized utterances collected from volunteer participants in Portal's in-house dogfooding program. The participants consist of households that have consented to having their Portal voice activity reviewed and analyzed. Every utterance in the evaluation set is associated with a personalized contact list, which we use for on-the-fly personalization to support calling queries. The contact list contains \textbf{876} names on average, with a standard deviation of \textbf{491}; this is several times larger than most of the bias lists used in \cite{Pundak2018DC,Jain2020DC}. The evaluation set is further split into three subsets:
\begin{itemize}
    \item \texttt{name-prod}: 4.7K utterances containing names from the personalized contact list.
    \item \texttt{name-rare}: 800 utterances containing names from the personalized contact list. The names in this subset are significantly more challenging than those typically seen in traffic and are used to benchmark our system on rare name recognition.
    \item \texttt{non-name}: 15.3K utterances without any name from the personalized contact list. Note that this subset still contains entities outside of the personalization data, such as artist names, city names, and song names.
\end{itemize}

We further augment our training data with TTS-generated utterances, motivated by previous successes of this technique in improving entity recognition \cite{He2019RNNT,Peyser2019,Zhao2019,Rosenberg19ASR+TTS}. We generate 1M transcripts for TTS through weighted sampling of 5.6K calling patterns mined from training data, combined with unweighted sampling of 710K name-pronunciation pairs from our internal pronunciation corpus. The transcripts are passed to our in-house phonetic TTS engine with neural vocoder, which verbalizes the input using four different voices (\{male, female\} $\times$ \{American English, British English\}), producing 4M utterances in total. Finally, we distort each utterance once using RIR and additive background noise prior to training.

\section{Methods}
\label{sec:methods}

\subsection{Shallow Fusion with Class-Based WFST LM}
\label{ssec:wfst_shallow_fusion}



\begin{figure}[tb]
  \centering
  \includegraphics[width=\columnwidth]{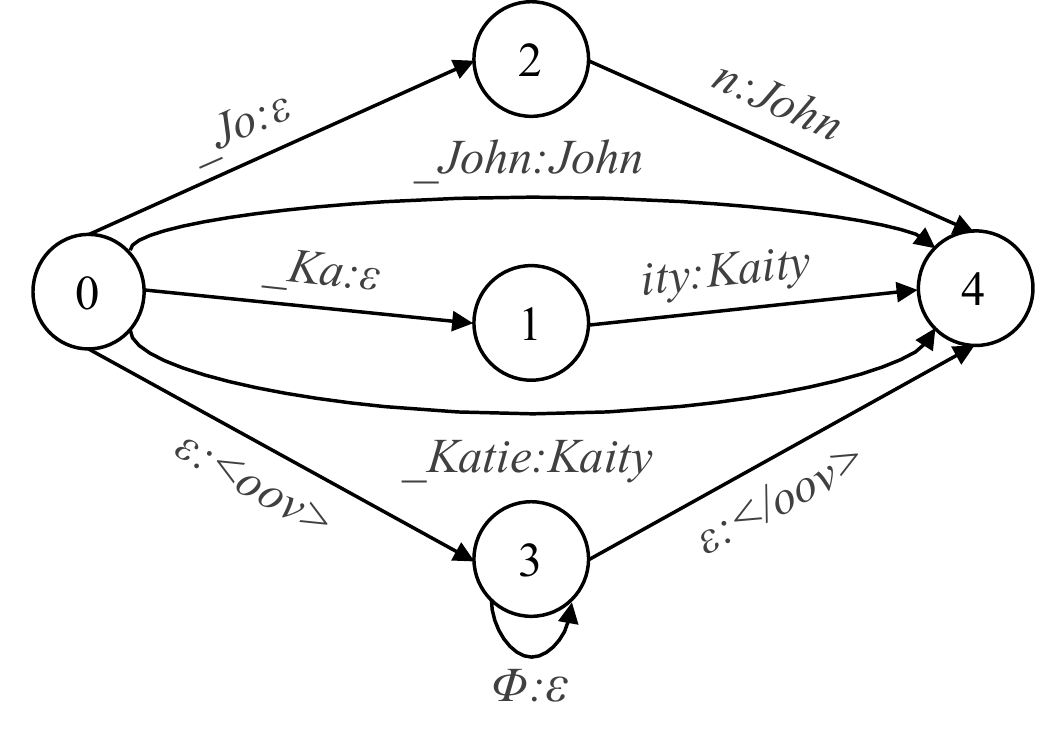}
  \caption{Example \texttt{@name} WFST with two names (John and Kaity), OOV failure arc, and one G2G pronunciation variant for each name (\texttt{\_Jo n} and \texttt{\_Katie}). The ilabel $\Phi$ indicates that any input symbol can be accepted by the OOV arc.}
  \label{fig:name_fst}
\end{figure}

We adopt a baseline WFST-based biasing approach for RNN-T in this work similar to \cite{Zhao2019}, where biasing is done at the WordPiece level and before the pruning stage of decoding. We employ a 4-gram WFST LM trained on a list of calling patterns mined from training data. The LM contains a special class tag, \texttt{@name}, which stands for the user's personalized contact list. Each word-level arc in the WFST is broken into WordPiece-level arcs according to SentencePiece \cite{kudo-richardson-2018-sentencepiece}, each of which has the same weight as the word-level arc. In decoding, the \texttt{@name} WFST is constructed dynamically from the personalized contact list, followed by determinization, minimization, and epsilon removal. To avoid over-biasing, we add an out-of-vocabulary (OOV) failure arc to the \texttt{@name} WFST to allow more flexible traversal through the class tag. This is similar to the failure arcs described in \cite{Zhao2019}; from early experiments we found that this OOV arc method produced better results. Figure \ref{fig:name_fst} shows an example \texttt{@name} WFST with two names, \emph{John} and \emph{Kaity}, along with the OOV failure arc.

\subsection{Sub-Word Regularization}
\label{ssec:subword_regularization}

The output units of RNN-T are typically WordPieces trained with Byte Pair Encoding (BPE) \cite{sennrich-etal-2016-neural} or unigram LM \cite{Kudo2018SubWord}. With large number of WordPieces, model training converges to word-level modeling as high-frequency words (e.g., ``hello" and ``interesting") are modeled whole. As a result, smaller WordPieces that make up these high-frequency words can be under-trained, leading to difficulties in recognizing rare names which are typically broken down into smaller WordPieces. Sub-word regularization was proposed in machine translation to alleviate this problem, where the reference WordPiece sequence is sampled from the n-best instead of taken from the best parse \cite{Kudo2018SubWord}. We hypothesize that a similar technique can be applied to RNN-T training and will be especially beneficial for name recognition. In addition to inducing better coverage of tail WordPieces, sub-word regularization can also help reduce model overconfidence and avoid early pruning. This technique has been used in ASR before \cite{Hannun19TDS}, but its impact on rare word recognition has not been studied.

\subsection{Encoder Pre-Training and Multi-Task Learning}
\label{ssec:pretrain_mtl}

The encoder, which can be viewed as RNN-T's built-in AM, is a key component of the model. Previous work has shown that pre-training the RNN-T encoder with Cross Entropy (CE) loss on frame-level force-aligned targets helps improve WER \cite{Hu2020PreTraining}. In this paper, we extend this technique by introducing an auxiliary CE loss on the encoder to predict frame-level targets, which is optimized jointly with the RNN-T loss. We use context- and position-dependent graphemes (i.e., chenones) \cite{Le2019Kulfi} as the target labels to avoid the use of phonetic lexicon. In this setting, multi-task learning (MTL) introduces complementary information and ensures consistent gradient flow into the encoder during training. We explore auxiliary training tasks for RNN-T in more detail in \cite{Liu2021MTL}.

\subsection{Leveraging G2G}
\label{ssec:leveraging_g2g}

RNN-T's output units are graphemic in nature, making it difficult to recognize rare names with poor grapheme-phoneme correspondence. In this work, we propose to leverage G2G \cite{Le2020G2G} to generate additional pronunciation variants for contact names during decoding. We decompose the output of G2G into WordPieces using the same SentencePiece model used by RNN-T while maintaining the word-level olabel. Figure \ref{fig:name_fst} shows an example \texttt{@name} WFST with G2G pronunciation variants added. With this method, the underlying WordPiece sequence produced by RNN-T does not have to match the word-level output, thus increasing the model's flexibility.

We also propose to leverage G2G during training by replacing each word in the reference text with a random G2G variant with some probability $p$. We do not replace words whose 1st-best G2G output is identical to their original written form, since these typically correspond to regular words that have consistent spelling and will likely not benefit from G2G replacement. Similar to sub-word regularization, G2G replacement helps increase WordPiece coverage during training and prevent model overconfidence.

\subsection{Deep PLM Fusion}
\label{ssec:wfst_deep_fusion}



%


\begin{figure}[tb]
  \centering
  \includegraphics[width=\columnwidth]{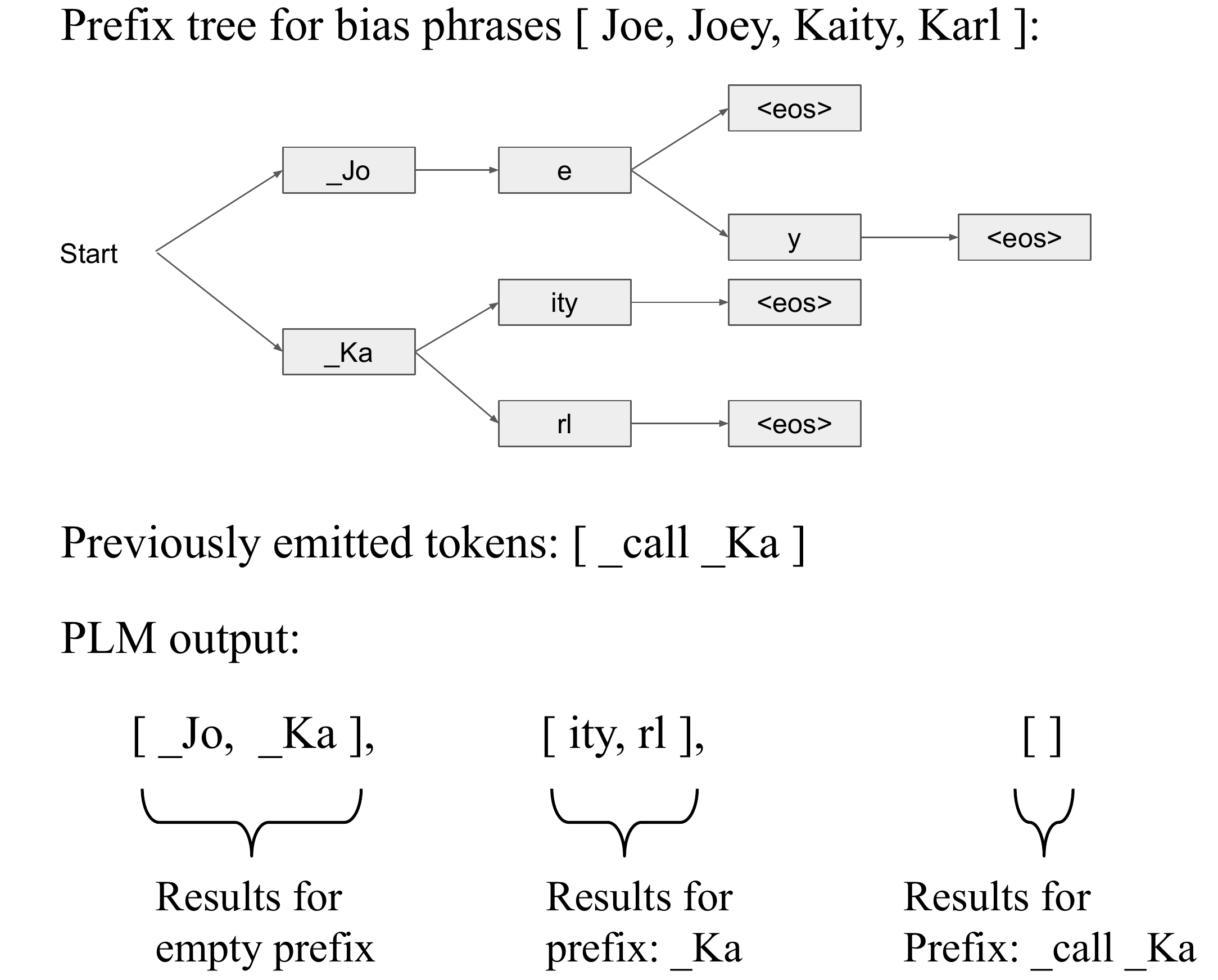}
  \caption{Example prefix tree and the results of query operations given a sequence of previously emitted symbols.}
  \label{fig:plm}
\end{figure}

One fundamental limitation of shallow fusion is that the combination with external LM scores happens after the RNN-T forward pass, during which the model does not have access to any information from the personalized contact list. Combined with RNN-T's spiky output, this lack of contextual information during the forward pass may limit RNN-T's ability to output the correct WordPiece sequence, especially for rare names. To address this issue, we introduce a component called \textbf{\emph{PLM predictor}}, which is responsible for transforming a sequence of previous output units $y_1, y_2, \ldots, y_{u-1}$ into an embedding vector $\mathbf{h}_u^{\text{plm}}$ given a list of contextual bias names $c_1, c_2, \ldots, c_N$ from the personalized contact list:
\begin{equation}
    \mathbf{h}^{\text{plm}}_u = f^{\text{plm}}(y_1, y_2, \ldots, y_{u-1}|c_1, c_2, \ldots, c_N)
\end{equation}

\noindent The joiner (Equation \ref{eq:joiner}) is then modified to estimate the logits $\mathbf{z}_{t,u}$ by taking into account the encoder, predictor, and PLM predictor embedding vectors:
\begin{equation}
    \mathbf{z}_{t,u} = f^{\text{join}}(\mathbf{h}_t^{\text{enc}} + \mathbf{h}_u^{\text{pred}} + \mathbf{h}^{\text{plm}}_u)
\label{eq:plm_joiner}
\end{equation}

There are many possible choices for the PLM predictor. We could design it as a NNLM that estimates the probability of the next WordPiece given previously emitted symbols. However, this requires adapting the NNLM on-the-fly given a contact list, which is difficult to do efficiently and effectively given relatively little data. We could re-use the WFST LM used in shallow fusion, where the set of valid ilabel transitions given the current FST states constitutes the embedding vector $\mathbf{h}_u^{\text{plm}}$. A challenge there is to make this operation efficient enough due to the presence of epsilon transitions and duplicate paths (the overall WFST is not determinized). To address this problem, we propose a simplified WFST LM called \textbf{\emph{simple PLM}} that can be queried more efficiently. This simple PLM builds a trie/prefix tree on a list of personalized contacts, each represented as a sequence of WordPieces. Given a length-$k$ prefix sequence of emitted symbols $y_{u-k}, \ldots, y_{u-1}$, the trie returns a list of valid next symbols represented as a binary vector $\mathbf{v}^u$ of length $V$, where $V$ is the number of WordPieces in the output vocabulary $\mathcal{Y}$:
\begin{equation}
    \text{Trie}(y_{u-k}, \ldots, y_{u-1}) =  \mathbf{v}^u
\end{equation}
where $\mathbf{v}^u_i = 1$ if $y_{u-k}, \ldots, y_{u-1}, \mathcal{Y}_i$ comprise a prefix of a contact name, and $0$ otherwise. Figure \ref{fig:plm} shows an example prefix tree and results of some query operations.

The PLM predictor output can then be defined as:
\begin{equation}
    \mathbf{h}_u^{\text{plm}} = W_{\text{plm}} [\text{Trie}(); \text{Trie}(y_{u-1}); \text{Trie}_{\ge 2}(y_1, \ldots, y_{u-1})]
\end{equation}

\noindent where $W_{\text{plm}}$ is a projection matrix to transform the concatenated binary vectors into the same dimension as the RNN-T encoder and predictor embeddings, and $\text{Trie}_{\ge 2}$ is the condensed binary vector for all prefixes of length $2$ or more:
\begin{equation}
\begin{split}
    \text{Trie}_{\ge 2}(y_1, \ldots, y_{u-1}) =& \hspace{0.25em} \text{Trie}(y_{u-2}, y_{u-1}) \hspace{0.25em} \texttt{OR} \\
    & \hspace{0.25em} \text{Trie}(y_{u-3}, y_{u-2}, y_{u-1}) \hspace{0.25em} \texttt{OR} \\
    & \hspace{0.25em} \ldots \\
    & \hspace{0.25em} \text{Trie}(y_1, y_2, \ldots, y_{u-1})
\end{split}
\end{equation}
\noindent where $\texttt{OR}$ stands for the element-wise logical OR operation. Intuitively, the first binary vector $\text{Trie}()$ represents symbols that can start a new contact name, the second binary vector $\text{Trie}(y_{u-1})$ represents symbols that can continue some contact name given the last emitted symbol as prefix, and the third binary vector $\text{Trie}_{\ge 2}(y_1, \ldots, y_{u-1})$ represents symbols that can continue some contact name given the last two or more emitted symbols as prefix.

We also experimented with attention-based PLM predictor similar to \cite{Pundak2018DC,Chen2019DC,Jain2020DC}, but found that this method did not scale well to large biasing lists and gave worse results while being significantly more expensive in training and inference. We therefore focus only on trie-based PLM predictor in this work.

\section{Experiments and Results}
\label{sec:results}

\subsection{Baseline System Results}
\label{ssec:baseline_system}

\begin{table}[t]
\centering
\begin{tabular}{ | c | E || D | D | D | }
    \hline
    \textbf{ID} & \textbf{Description} & \texttt{name-prod} & \texttt{name-rare} & \texttt{non-name} \\
    \hline
    \multicolumn{5}{l}{} \\
    \multicolumn{5}{l}{\emph{1. Baseline Systems}} \\
    \hline
    \texttt{H1} & Hybrid & 10.6 & 9.3 & 7.8 \\
    \texttt{B1} & RNN-T & 33.5 & 42.0 & 7.0 \\
    \textbf{\texttt{B2}} & \textbf{\texttt{B1} + Shallow Fusion} & \textbf{11.7} & \textbf{13.9} & \textbf{7.9} \\
    \hline
    \multicolumn{5}{l}{} \\
    \multicolumn{5}{l}{\emph{2. With Sub-Word Regularization}} \\
    \multicolumn{5}{l}{\emph{($l$: sampling size, $\alpha$: smoothing parameter \cite{Kudo2018SubWord})}} \\
    \hline
    \textbf{\texttt{S1}} & \textbf{\texttt{B2} + ($\mathbf{l}$=5, $\bm{\alpha}$=0.25)} & \textbf{10.6} & \textbf{12.9} & \textbf{8.6} \\
    \texttt{S2} & \texttt{B2} + ($l$=10, $\alpha$=0.25) & 10.5 & 14.5 & 7.9 \\
    \texttt{S3} & \texttt{B2} + ($l$=15, $\alpha$=0.25) & 11.3 & 13.9 & 8.3 \\
    \hline
    \multicolumn{5}{l}{} \\
    \multicolumn{5}{l}{\emph{3. With Pre-Training (PT) and Multi-Task Learning (MTL)}} \\
    \hline
    \texttt{M1} & \texttt{S1} + PT & 10.2 & 11.5 & 9.5 \\
    \texttt{M2} & \texttt{S1} + MTL & 10.7 & 13.3 & 9.0 \\
    \textbf{\texttt{M3}} & \textbf{\texttt{S1} + PT and MTL} & \textbf{10.2} & \textbf{12.3} & \textbf{8.6} \\
    \hline
    \multicolumn{5}{l}{} \\
    \multicolumn{5}{l}{\emph{4. With G2G in Training and/or Decoding}} \\
    \hline
    \texttt{G1} & \texttt{M3} + G2G Train & 9.8 & 11.7 & 8.5 \\
    \texttt{G2} & \texttt{M3} + G2G Decode & 10.9 & 10.8 & 8.9 \\
    \textbf{\texttt{G3}} & \textbf{\texttt{M3} + G2G All} & \textbf{10.1} & \textbf{10.2} & \textbf{8.4} \\
    \hline
    \multicolumn{5}{l}{} \\
    \multicolumn{5}{l}{\emph{5. With Deep PLM Fusion}} \\
    \multicolumn{5}{l}{\emph{(frozen encoder and predictor)}} \\
    \hline
    \texttt{P1} & \texttt{G3} + Deep PLM & 10.3 & 9.3 & 8.4 \\
    \texttt{P2} & \texttt{P1} + Sub-Word Reg. & 10.1 & 9.3 & 8.4 \\
    \textbf{\texttt{P3}} & \textbf{\texttt{P2} + G2G Train} & \textbf{9.9} & \textbf{9.1} & \textbf{8.3} \\
    \hline
\end{tabular}
\caption{Word Error Rate summary of proposed techniques.}
\label{table:overall_results}
\end{table}

The baseline RNN-T used in this work has ~37M parameters in total. The encoder takes 11 stacked logMel feature frames as input, has eight layers of Long-Short Term Memory (LSTM) with 640 units, and subsamples the input by a factor of four. The predictor consists of two LSTM layers with 256 units. We apply 0.3 dropout and layer normalization between every layer in both the encoder and predictor, and their outputs are projected into 1024 dimensions using a fully-connected (FC) layer. Finally, the joiner contains one FC layer with input dimension 1024. The output targets are 4096 unigram WordPiece units \cite{Kudo2018SubWord} trained with SentencePiece \cite{kudo-richardson-2018-sentencepiece}. The model is trained for 50 epochs with Adam optimizer \cite{kingma2014adam}; the learning rate is fixed at 0.0002 for the first 35 epochs, then decays by 0.85 after every epoch until the end of training. The epoch with the best validation loss is chosen for evaluation.

Table \ref{table:overall_results} (Section 1) shows that, as expected, shallow fusion improves name recognition significantly while incurring some degradation on \texttt{non-name}. Compared to a hybrid system consisting of a LSTM AM trained on chenones and a large 5-gram word-level LM (several GBs), RNN-T with shallow fusion performs similarly on \texttt{non-name}, but underperforms by 10.4\% relative on \texttt{name-prod} and 49.5\% on \texttt{name-rare}. These results indicate RNN-T's problems with entity recognition, especially on rare names.
 
\subsection{Effect of Sub-Word Regularization}
\label{ssec:subword_regularization_results}

Table \ref{table:overall_results} (Section 2) shows RNN-T results after adding sub-word regularization with fixed smoothing parameter $\alpha = 0.25$ and different sampling sizes $l$. Excessively large $l$ leads to unstable training, and the best result was achieved with $l=5$. In this setup, we obtain relative improvement of 9.4\% on \texttt{name-prod} and 7.2\% on \texttt{name-rare}, while degrading \texttt{non-name} by 8.9\%. The degradation on \texttt{non-name} could possibly be avoided by performing sub-word regularization only on entity names, which we will explore in future work.

\subsection{Effect of Pre-Training and Multi-Task Learning}
\label{ssec:pretrain_mtl_results}

We pre-train the RNN-T encoder for 20 epochs with CE loss on chenone targets. During RNN-T training, the auxiliary CE loss is attached to the last encoder layer and optimized jointly with the RNN-T loss (the CE loss weight is set at 0.1). Table \ref{table:overall_results} (Section 3) shows the impact of these techniques. Combining pre-training and MTL gave the best overall performance, resulting in an improvement of 3.8\% on \texttt{name-prod}, 4.7\% on \texttt{name-rare}, and no change on \texttt{non-name} compared to the \texttt{S1} baseline. We could likely achieve more gain by tuning the location of the auxiliary loss as well as the CE loss weight. This will be investigated in a follow-up work.

\subsection{Effect of G2G}
\label{ssec:g2g_results}

The G2G model used in this work follows the same training procedure as in \cite{Le2020G2G}, but uses RNN-T instead of hybrid ASR to re-decode the TTS output. We generate two additional G2G variants for each word in addition to the identity mapping, and fix the replacement probability $p$ during training to 0.2. As seen in Table \ref{table:overall_results} (Section 4), applying G2G only to training leads to small but consistent improvement across all test sets. Applying G2G only to decoding leads to significant improvement on \texttt{name-rare}, but degrades the other two test sets due to increased confusion in the WFST. The problem is mitigated by applying G2G to both training and decoding, where we see similar performance on \texttt{name-prod}, significant 17.1\% improvement on \texttt{name-rare}, and slight 2.3\% improvement on \texttt{non-name} compared to the \texttt{M3} baseline.

\subsection{Effect of Deep PLM Fusion}
\label{ssec:deep_plm_fusion_results}

\begin{table}[t]
\centering
\begin{tabular}{ | F | F || D | D | D | }
    \hline
    \textbf{Shallow Fusion} & \textbf{G2G (Decoding)} & \texttt{name-prod} & \texttt{name-rare} & \texttt{non-name} \\
    \hline
    \hline
    \texttt{NO} & \texttt{NO} & 25.7 & 30.6 & 8.2 \\
    \texttt{YES} & \texttt{NO} & 10.0 & 9.9 & 8.3 \\
    \textbf{\texttt{YES}} & \textbf{\texttt{YES}} & \textbf{9.9} & \textbf{9.1} & \textbf{8.3} \\
    \hline
\end{tabular}
\caption{Impact of shallow fusion and G2G on deep PLM.}
\label{table:deep_plm_fusion_ablation}
\end{table}

Unlike test data, our training utterances are not associated with any personalized contact list and the target entity (if any) is unknown; both are required for deep PLM training. We identify potential target entities in each training utterance using an in-house entity tagger. We then simulate a contact list for each training utterance on-the-fly by sampling between 200 and 400 names from the list of tagged entities. We further expand this contact list by adding two G2G variants for each name, which makes the simulated contact list more realistic. To speed up training, we initialize and freeze the RNN-T encoder and predictor weights from the \texttt{G3} baseline, thus only the joiner and PLM predictor are trained from scratch. Freezing the encoder and predictor also avoids spurious WER improvement due to training the model longer. We train the model for 20 epochs with Adam; the learning rate is fixed at 0.0001 for the first 10 epochs, then decays by 0.5 every epoch.

In initial experiments, we found that RNN-T with deep PLM is prone to overfit due to an over-reliance on the PLM predictor. The model's n-best list tends to contain very diverse contact names rather than acoustically similar words. This poor confusion pattern especially hurts the model when combined with shallow fusion. We henceforth propose three regularization techniques to combat this issue. First, we remove the target entity from the simulated contact list with 0.5 probability. Second, we apply 0.3 dropout to the PLM predictor output $\mathbf{h}_u^{\text{plm}}$. Third, inspired by scheduled sampling \cite{Bengio2015scheduled}, we replace the target entity with another name with 0.3 probability before feeding the reference text through the predictors. All three techniques force the model to rely more on the encoder and less on the predictors during training. Although these regularization methods slightly degraded WER of the stand-alone deep PLM model, we found they were crucial for getting additional improvement on top of shallow fusion.

Our proposed techniques, together with sub-word regularization and G2G replacement, allow deep PLM fusion to improve upon the \texttt{G3} baseline, as seen in Table \ref{table:overall_results} (Section 5). Deep PLM is especially helpful for \texttt{name-rare}, yielding 10.8\% relative improvement. The benefit will likely be higher with a weaker but more generic WFST LM that does not rely as much on known calling patterns. Compared to the RNN-T baseline \texttt{B2}, our final system improves over \texttt{name-prod} by \textbf{15.4\%} and \texttt{name-rare} by \textbf{34.5\%}, while degrading \texttt{non-name} by 5.1\%. Interestingly, the gap on name recognition with the hybrid baseline \texttt{H1} has been closed completely.

Lastly, the ablation study in Table \ref{table:deep_plm_fusion_ablation} shows that deep PLM fusion in its current form cannot completely replace shallow fusion, despite clear improvement over the baseline \texttt{B1}. We will investigate the root cause of this performance gap in future work. In addition, the last two rows of the table indicate that G2G and deep PLM fusion provide complementary gains.

\section{Conclusion and Future Work}
\label{sec:conclusion}


In this paper, we showed that RNN-T personalization can be improved significantly by inducing better coverage of rare WordPieces during training, introducing extra information into the encoder, leveraging G2G to produce additional pronunciation variants in both training and decoding, and biasing earlier with deep PLM fusion. Together, these techniques help push the boundary of RNN-T personalization and close the gap with traditional hybrid systems on use cases that require contextual biasing and accurate name recognition. For future work, we plan to incorporate proper WFST and NNLM into deep PLM fusion, apply these techniques to other end-to-end models, and tackle open-domain personalization where strong context prefixes are not always available.

\section{ACKNOWLEDGMENT}
\label{sec:acknowledgment}

We'd like to thank Mahaveer Jain, Jiedan Zhu, and Xuedong Zhang for their helpful advice and suggestions.

\newpage
\small
\bibliographystyle{IEEEbib}
\bibliography{refs}

\begin{thebibliography}{10}

\bibitem{Chorowski:2015:AMS:2969239.2969304}
J.~Chorowski, D.~Bahdanau, D.~Serdyuk, K.~Cho, and Y.~Bengio,
\newblock ``{Attention-based Models for Speech Recognition},''
\newblock in {\em Proc. NIPS}, 2015.

\bibitem{chan2015listen}
W.~{Chan}, N.~{Jaitly}, Q.~{Le}, and O.~{Vinyals},
\newblock ``{Listen, attend and spell: A neural network for large vocabulary
  conversational speech recognition},''
\newblock in {\em Proc. ICASSP}, 2016.

\bibitem{Chiu18}
C.~{Chiu}, T.~N. {Sainath}, Y.~{Wu}, R.~{Prabhavalkar}, P.~{Nguyen}, Z.~{Chen},
  A.~{Kannan}, R.~J. {Weiss}, K.~{Rao}, E.~{Gonina}, N.~{Jaitly}, B.~{Li},
  J.~{Chorowski}, and M.~{Bacchiani},
\newblock ``{State-of-the-Art Speech Recognition with Sequence-to-Sequence
  Models},''
\newblock in {\em Proc. ICASSP}, 2018.

\bibitem{zeyer2018improved}
A.~Zeyer, K.~Irie, R.~Schl{\"u}ter, and H.~Ney,
\newblock ``Improved training of end-to-end attention models for speech
  recognition,''
\newblock in {\em Proc. INTERSPEECH}, 2018.

\bibitem{Graves12transduction}
A.~Graves,
\newblock ``Sequence transduction with recurrent neural networks,''
\newblock in {\em ICML Representation Learning Workshop}, 2012.

\bibitem{Prabhavalkar17}
R.~Prabhavalkar, K.~Rao, T.~Sainath, B.~Li, L.~Johnson, and N.~Jaitly,
\newblock ``{A Comparison of Sequence-to-Sequence Models for Speech
  Recognition},''
\newblock in {\em Proc. INTERSPEECH}, 2017, pp. 939--943.

\bibitem{Battenberg17RNNT}
E.~{Battenberg}, J.~{Chen}, R.~{Child}, A.~{Coates}, Y.~G.~Y. {Li}, H.~{Liu},
  S.~{Satheesh}, A.~{Sriram}, and Z.~{Zhu},
\newblock ``Exploring neural transducers for end-to-end speech recognition,''
\newblock in {\em Proc. ASRU}, 2017.

\bibitem{He2019RNNT}
Y.~{He}, T.~N. {Sainath}, R.~{Prabhavalkar}, I.~{McGraw}, R.~{Alvarez},
  D.~{Zhao}, D.~{Rybach}, A.~{Kannan}, Y.~{Wu}, R.~{Pang}, Q.~{Liang},
  D.~{Bhatia}, Y.~{Shangguan}, B.~{Li}, G.~{Pundak}, K.~C. {Sim}, T.~{Bagby},
  S.~{Chang}, K.~{Rao}, and A.~{Gruenstein},
\newblock ``{Streaming End-to-end Speech Recognition for Mobile Devices},''
\newblock in {\em Proc. ICASSP}, 2019.

\bibitem{Kannan2018}
A.~Kannan, Y.~Wu, P.~Nguyen, T.~N. Sainath, Z.~Chen, and R.~Prabhavalkar,
\newblock ``{An Analysis of Incorporating an External Language Model Into a
  Sequence-to-Sequence Model},''
\newblock in {\em Proc. ICASSP}, 2018.

\bibitem{Zhao2019}
D.~Zhao, T.~N. Sainath, D.~Rybach, P.~Rondon, D.~Bhatia, B.~Li, and R.~Pang,
\newblock ``{Shallow-Fusion End-to-End Contextual Biasing},''
\newblock in {\em Proc. INTERSPEECH}, 2019.

\bibitem{Pundak2018DC}
G.~Pundak, T.~N. Sainath, R.~Prabhavalkar, A.~Kannan, and D.~Zhao,
\newblock ``{Deep Context: End-to-End Contextual Speech Recognition},''
\newblock in {\em Proc. ICASSP}, 2018.

\bibitem{Chen2019DC}
Z.~Chen, M.~Jain, Y.~Wang, M.~L. Seltzer, and C.~Fuegen,
\newblock ``{Joint Grapheme and Phoneme Embeddings for Contextual End-to-End
  ASR},''
\newblock in {\em Proc. INTERSPEECH}, 2019.

\bibitem{Jain2020DC}
M.~Jain, G.~Keren, J.~Mahadeokar, G.~Zweig, F.~Metze, and Y.~Saraf,
\newblock ``{Contextual RNN-T For Open Domain ASR},''
\newblock in {\em Proc. INTERSPEECH}, 2020.

\bibitem{Irie2019output}
K.~Irie, R.~Prabhavalkar, A.~Kannan, A.~Bruguier, D.~Rybach, and P.~Nguyen,
\newblock ``{Model Unit Exploration for Sequence-to-Sequence Speech
  Recognition},''
\newblock in {\em Proc. INTERSPEECH}, 2019.

\bibitem{Peyser2019}
C.~Peyser, H.~Zhang, T.~N. Sainath, and Z.~Wu,
\newblock ``{Improving Performance of End-to-End ASR on Numeric Sequences},''
\newblock in {\em Proc. INTERSPEECH}, 2019.

\bibitem{Rosenberg19ASR+TTS}
A.~Rosenberg, Y.~Zhang, B.~Ramabhadran, Y.~Jia, P.~Moreno, Y.~Wu, and Z.~Wu,
\newblock ``{Speech Recognition with Augmented Synthesized Speech},''
\newblock in {\em Proc. ASRU}, 2019.

\bibitem{Kudo2018SubWord}
T.~Kudo,
\newblock ``{Subword Regularization: Improving Neural Network Translation
  Models with Multiple Subword Candidates},''
\newblock in {\em Proc. ACL}, 2018.

\bibitem{Hu2020PreTraining}
H.~Hu, R.~Zhao, J.~Li, L.~Lu, and Y.~Gong,
\newblock ``{Exploring Pre-training with Alignments for RNN Transducer based
  End-to-End Speech Recognition},''
\newblock in {\em Proc. ICASSP}, 2020.

\bibitem{Liu2021MTL}
C.~Liu, F.~Zhang, D.~Le, S.~Kim, Y.~Saraf, and G.~Zweig,
\newblock ``{Improving RNN Transducer Based ASR with Auxiliary Tasks},''
\newblock in {\em Proc. SLT}, 2021.

\bibitem{Le2020G2G}
D.~Le, T.~Koehler, C.~Fuegen, and M.~L. Seltzer,
\newblock ``{G2G: TTS-Driven Pronunciation Learning for Graphemic Hybrid
  ASR},''
\newblock in {\em Proc. ICASSP}, 2020.

\bibitem{chiu2019comparison}
C.~C. Chiu, W.~Han, Y.~Zhang, R.~Pang, S.~Kishchenko, P.~Nguyen, A.~Narayanan,
  H.~Liao, S.~Zhang, A.~Kannan, et~al.,
\newblock ``{A Comparison of End-to-End Models for Long-Form Speech
  Recognition},''
\newblock in {\em Proc. ASRU}, 2019.

\bibitem{li2020comparison}
J.~Li, Y.~Wu, Y.~Gaur, C.~Wang, R.~Zhao, and S.~Liu,
\newblock ``{On the Comparison of Popular End-to-End Models for Large Scale
  Speech Recognition},''
\newblock in {\em Proc. INTERSPEECH}, 2020.

\bibitem{gulcehre2015using}
C.~Gulcehre, O.~Firat, K.~Xu, K.~Cho, L.~Barrault, H.~C. Lin, F.~Bougares,
  H.~Schwenk, and Y.~Bengio,
\newblock ``On using monolingual corpora in neural machine translation,''
\newblock {\em arXiv preprint arXiv:1503.03535}, 2015.

\bibitem{sriram2018cold}
A.~Sriram, H.~Jun, S.~Satheesh, and A.~Coates,
\newblock ``{Cold Fusion: Training Seq2Seq Models Together with Language
  Models},''
\newblock in {\em Proc. INTERSPEECH}, 2018.

\bibitem{Ko2015AudioAF}
T.~Ko, V.~Peddinti, D.~Povey, and S.~Khudanpur,
\newblock ``Audio augmentation for speech recognition,''
\newblock in {\em Proc. INTERSPEECH}, 2015.

\bibitem{kudo-richardson-2018-sentencepiece}
T.~Kudo and J.~Richardson,
\newblock ``{{S}entence{P}iece: A simple and language independent subword
  tokenizer and detokenizer for Neural Text Processing},''
\newblock in {\em Proc. EMNLP: System Demonstrations}, 2018.

\bibitem{sennrich-etal-2016-neural}
R.~Sennrich, B.~Haddow, and A.~Birch,
\newblock ``{Neural Machine Translation of Rare Words with Subword Units},''
\newblock in {\em Proc. ACL}, 2016.

\bibitem{Hannun19TDS}
A.~Hannun, A.~Lee, Q.~Xu, and R.~Collobert,
\newblock ``{Sequence-to-Sequence Speech Recognition with Time-Depth Separable
  Convolutions},''
\newblock in {\em Proc. INTERSPEECH}, 2019.

\bibitem{Le2019Kulfi}
D.~Le, X.~Zhang, W.~Zheng, C.~Fuegen, G.~Zweig, and M.~L. Seltzer,
\newblock ``{From Senones to Chenones: Tied Context-Dependent Graphemes for
  Hybrid Speech Recognition},''
\newblock in {\em Proc. ASRU}, 2019.

\bibitem{kingma2014adam}
D.~P. Kingma and J.~Ba,
\newblock ``{Adam: A Method for Stochastic Optimization},''
\newblock in {\em Proc. ICLR}, 2014.

\bibitem{Bengio2015scheduled}
S.~Bengio, O.~Vinyals, N.~Jaitly, and N.~Shazeer,
\newblock ``{Scheduled Sampling for Sequence Prediction with Recurrent Neural
  Networks},''
\newblock in {\em Proc. NIPS}, 2015.

\end{thebibliography}

\end{document}